\documentclass[final,5p,times,twocolumn]{elsarticle}
\usepackage{amsmath,amssymb,amsfonts}
\usepackage{algorithmic}
\usepackage{graphicx}
\usepackage{textcomp}
\usepackage{float}
\usepackage{balance,flushend}
\usepackage{xcolor}

 \usepackage{threeparttable}

\usepackage{amssymb}

 \journal{Journal of Guangdong University of Technology}

\begin{document}

\begin{frontmatter}

\title{Active Mining Sample Pair Semantics for Image-text Matching}

\author{Yongfeng Chen
\fnref{a,fn}}
\ead{2112103118@mail2.gdut.edu.cn}

\author{Jin Liu
\fnref{a,fn}}
\ead{2111903101@mail2.gdut.edu.cn}

\author{Zhijing Yang
\fnref{a}}
\ead{yzhj@gdut.edu.cn}

\author{Ruihan Chen
\fnref{a}}
\ead{2112103075@mail2.gdut.edu.cn}

\author{Junpeng Tan
\fnref{b}}
\ead{tjeep@mail2.gdut.edu.cn}

\cortext[cor]{Corresponding author (e-mail:yzhj@gdut.edu.cn)}

\affiliation[a]{organization={School of Information Engineering, Guangdong University of Technology},
city={Guangzhou},
postcode={510006}, 
country={China}}
            
\affiliation[b]{organization={School of Electronic and Information Engineering, South China University of Technology},
city={Guangzhou},
postcode={510641}, 
country={China}}

\fntext[fn]{The first two authors share equal contributions.}









            
            

\begin{abstract}
Recently, commonsense learning has been a hot topic in image-text matching. Although it can describe more graphic correlations, commonsense learning still has some shortcomings: 1) The existing methods are based on triplet semantic similarity measurement loss, which cannot effectively match the intractable negative in image-text sample pairs. 2) The weak generalization ability of the model leads to the poor effect of image and text matching on large-scale datasets. According to these shortcomings. This paper proposes a novel image-text matching model, called Active Mining Sample Pair Semantics image-text matching model (AMSPS). Compared with the single semantic learning mode of the commonsense learning model with triplet loss function, AMSPS is an active learning idea. Firstly, the proposed Adaptive Hierarchical Reinforcement Loss (AHRL) has diversified learning modes. Its active learning mode enables the model to more focus on the intractable negative samples to enhance the discriminating ability. In addition, AMSPS can also adaptively mine more hidden relevant semantic representations from uncommented items, which greatly improves the performance and generalization ability of the model. Experimental results on Flickr30K and MSCOCO universal datasets show that our proposed method is superior to advanced comparison methods.

\end{abstract}


\end{frontmatter}



\section{Introduction}
Image-Text Matching (ITM) \cite{lee2018stacked,li2019visual,beltran2021deep}, is at the forefront of multi-modal or cross-modal fields. ITM refers to taking a given text or image as the source to retrieve another modal object with the most similar semantic information. Technically, ITM needs to map information such as image and text features into the same semantic space, and then judge the matching degree by similarity measurement \cite{szegedy2016rethinking}.

Noticeably, deep semantic learning networks \cite{lu2020infrared,sun2022superpoint,lu2020robust} have achieved great progress and attention in visual-semantic embedding \cite{li2019visual,faghri2017improving}. For example, Li et al. \cite{li2019visual} used the Graph Convolutional Networks (GCN) method to infer the relationship between objects in images to improve the performance of image-text matching. However, the limitation of the visual-semantic embedding method is that it can only discover semantic concepts from the image-modal, and cannot mine the potential close relationship between words in the text-modal. Such as {“beach”, “sea” and “boat”}, these three words have a strong internal connection. When one word appears, there is a high probability that the other two words will appear together. This commonsense concept can obtain higher similarity by using the collaborative text connection relational network model.

Moreover, the existing visual-semantic ITM methods are based on the triplet loss function to measure cross-modal similarity \cite{faghri2017improving,tan2023cross,chen2024asymmetric}. They usually take small batch samples of an anchor for training. The pairs of samples that can be matched are called positive pairs, and negative pairs else. The triplet loss function forces positive pairs to cluster and negative pairs to separate in the embedding space. However, the selected negative pairs \cite{wang2017deep} of the triplet loss are still not accurate enough. The ranking of negative pairs of an anchor is usually very low. The ability of this model to distinguish positive pairs and negative pairs is still not enough for the retrieval effect of complex image and text information. Furthermore, the triplet loss can only be limited to their own samples, and positive/negative pairs of other samples cannot be introduced as a guide during training. This defect inevitably leads to the weak generalization ability of the model \cite{tan2022novel}. Finally, another disadvantage of the triplet loss function is that its boundary is fixed, which does not meet the current trend of big data information retrieval.

To address these challenges, in this paper, we choose the commonsense learning model, Consensus-aware Visual-Semantic Embedding (CVSE) \cite{wang2020consensus} as our backbone framework. Inspired by active learning \cite{boukthir2022reduced,gweon2021nearest}, we further add predictive candidates into the original image and text features to help model training. Besides, we propose Adaptive Hierarchical Reinforcement Loss (AHRL) based on the Consensus-based Image Description Evaluation (CIDE) \cite{vedantam2015cider} metric to adaptively learn optimal image-text pairs. To this end, we propose a novel network model that Active Mining Sample Pair Semantics (AMSPS) for the ITM task. The main contributions can be listed as follows:
\begin{enumerate}
\item We introduce two prediction candidates, which include a small number of intractable negative pairs, and build penalties to better guide the model on how to separate negative pairs. Moreover, we employ the hybrid training method to make up for the weak generalization ability of the triplet loss function by the increase of sample diversity.

\item We propose an AHSL based on the CIDE metric, which can not only better learn the semantics of indistinguishable sample pairs, but also capture the relevant semantics of non-ground-truth items well, greatly improving the score ranking and the generalization ability of the model.

\item Extensive experiments on the two benchmark datasets prove the superiority of our method over previous work. The corresponding ablation study demonstrates the rationality and effectiveness of each component in our method.
\end{enumerate}

\begin{figure*}[htb]
    \centering
    \includegraphics[scale=0.243]{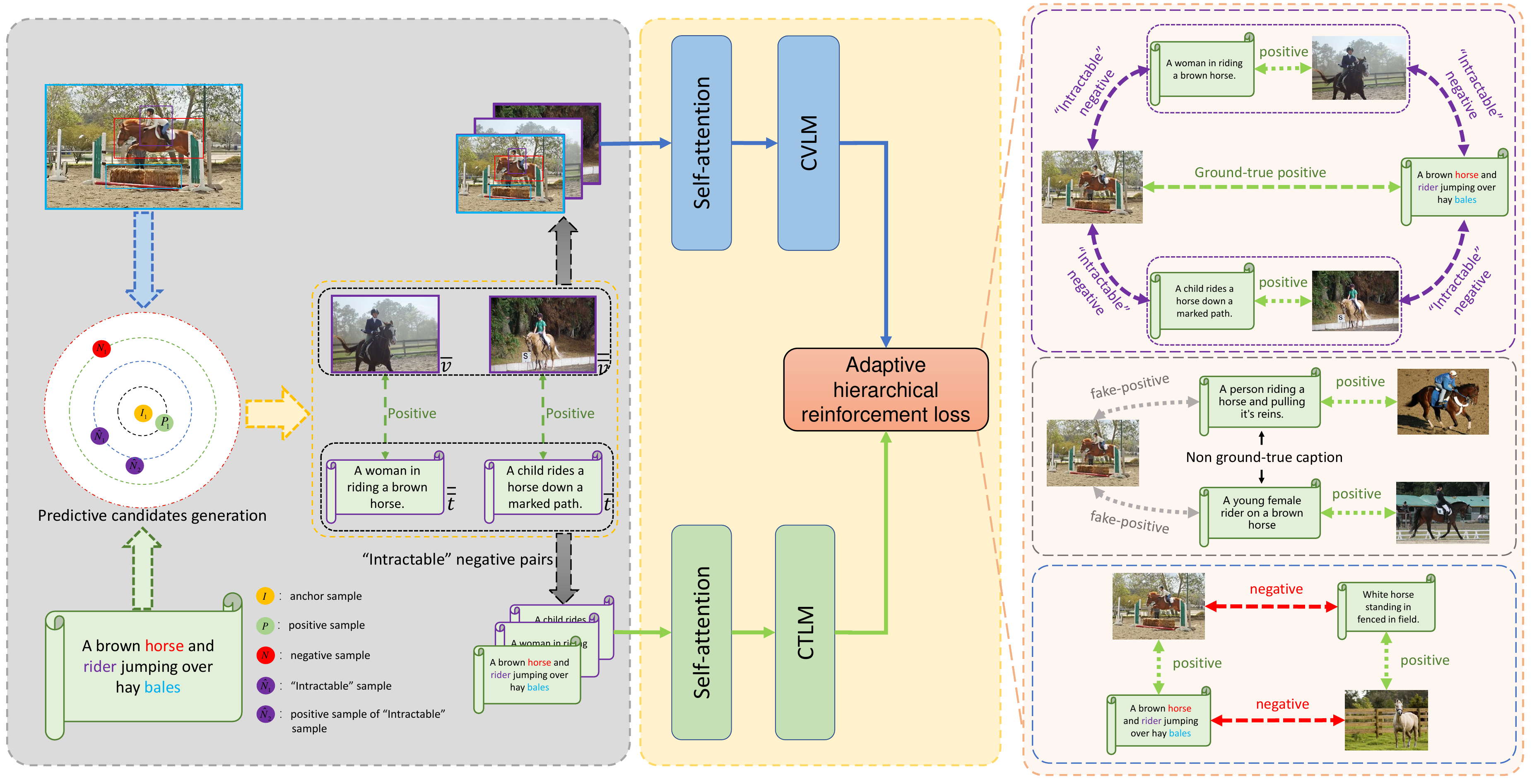}
    \caption{An overview of the proposed AMSPS for ITM. The left: (a) extracting and encoding images and texts as input representation, and adding additional predictive candidates for collaborative training. The middle: (b) The global representations of images and texts are obtained by semantic attention modules CVTM and CTLM, and finally aligned globally using AHRL. The right: (c) AHRL ($L_{AHRL}$) will study the information of different sample pairs as much as possible. AHSL can learn "intractable" negative pairs to enhance the discriminant ability of the model to distinguish negative pairs. It can also mine and learn more additional semantic information related to images from non-ground-truth captions.}
    \label{fig:ITMframe}
\end{figure*}

\section{Related work}

\subsection{ITM Methods}

Global or local embedding techniques are commonly used in ITM tasks. The global alignment techniques \cite{matsubara2021target,huang2023context} generally embed the entire image or complete sentence into the common space to calculate its similarity. Matsubara et al. \cite{matsubara2021target} proposed a target-oriented deformation network that can constantly learn from embedded spaces into new spaces by adjusting the similarity of global image-text pairs. However, global alignment techniques cannot accurately capture the details of cross-modal samples. Therefore, most people focus on local alignment methods \cite{karpathy2015deep}. 

The local alignment methods can explore the fine-grained relationship between image-text pairs. Karpathy et al. \cite{karpathy2015deep} proposed a structure that can infer the potential alignment between text words and image areas, and connect image features with text features through a common embedded space. However, these ITM methods are limited to global and local alignment learning using image-text instance pairs. Cross-modal semantic concepts can only be detected from specific images.

On the other hand, global and local alignment learning techniques make dealing with long tails and occlusion concepts difficult. Therefore, an advanced method was proposed, called commonsense learning, which uses the frequently appearing co-occurrence concepts in the model as commonsense concepts to guide the model and then can better learn visual-semantic. Wang et al. \cite{wang2020consensus} used their own corpora for external embedding. By calculating the correlation between semantic concepts of the corpus, the concept correlation graph is constructed. The consensus representation is then obtained and embedded into the model. Although these methods can learn the cross-modal correlation of image and text,  there are intractable negative pairs for some complex image-text pairs.

\subsection{ITM Losses}

The similarity loss measure is also an important direction for optimizing the ITM model. Early ITM model training is mainly based on the triplet loss function. The triplet loss function excites the model well to narrow the distance between the anchor point and the positive instance pairs. Recently, Liu et al. \cite{liu2019neighbor} proposed a neighbor-aware sorting loss method, which focuses on neighbors belonging to different semantics and effectively distinguishes different semantics. Liu et al. [34] proposed hubness-aware loss, which considers all samples and uses local and global statistics to expand the weight of the hub.

Different from the above works, we will focus on deep semantic mining of inter- and intra-modal samples. Here, we present an AHRL. The intractable negative pairs can be mined to improve the ability of the model to distinguish negative pairs. Adaptively learning the semantic information of non-ground-truth captions greatly improves the model matching score and generalization ability.

\section{Methodology}
In this section, we introduce our proposed ITM framework AMSPS and loss measure AHRL. As shown in Fig. 1,  firstly, we introduce distinct input candidates in section \ref{AA}. Secondly, we describe the works of the Corpus Visual Learning Module (CVLM) and Corpus Textual Learning Module (CTLM) in section \ref{BB}. Finally, we focus on introducing our proposed AHRL in the section \ref{cc}.

\subsection{The input Candidates}\label{AA}
\textbf{Image Candidates.} To deal with the images with complex content, we introduce image instance candidates. In this way, we can obtain more semantic information and correlation between objects in the image. Firstly, we use the pre-trained bottom-up attention mechanism \cite{anderson2018bottom} with a Faster R-CNN \cite{ren2015faster} model to process the images. Specifically, given an image $I$, we use a pre-trained model to detect each region $o$ in the image and extract the feature vector which is denoted as $f_{o}$ with 2048-dimensional from each region. In Eq. (1), we transform $f_{o}$ into a low-dimensional vector $v_{o}$ by adding a fully connected layer:
\begin{equation}
v_{o}=W_{f}f_{o}+b_{f}
\end{equation}
where $W_{f}$ and $b_{f}$ are the network parameters that need to be learned. Each image is represented by the set $V=\{v_{1},...,v_{o}\}, V \in R^{D\times o}$ which is the image candidates, where $D$ is the dimensionality of $v_o$. Finally, we normalize each feature $V$, like Ref. \cite{lee2018stacked}.

\textbf{Text Candidates}. Given a sentence $T$ with $s$ words $\{w_{1},...,w_{s}\}$, we use the common embedding matrix $W_e$ to link each word as a 300-dimensional word embedding vector as $e_{j}=W_{e}w_{j},j\in [1,s]$. To further strengthen the association between words, we need to use a structural framework that can both relate past context information and predict future context information, e.g., bi-directional GRU \cite{schuster1997bidirectional}. The formula is as follows:
\begin{equation}
{\overrightarrow{h_j^f}=\overrightarrow{GRU^{f}}(e_{j}+h_{j-1}^f) ; \overleftarrow{h_j^b}=\overleftarrow{GRU^{b}}(e_{j}+h_{j-1}^b)}
\end{equation}
where $\overrightarrow{h_j^f}$ and $\overleftarrow{h_j^b}$ denote hidden states that mean forward pass and backward pass at time step $j$, respectively. Then we average the two hidden states as $t_j^a=\frac{\overrightarrow{h_j^f}+\overleftarrow{h_j^b}}{2}$, which can obtain two different contextual information centered on $e_j$. Eventually, the global textual feature vectors are represented by $T=\{t_j^a\mid j=1,...,n, t_j^a\in R^d\}$. Similarly, $T$ is normalized in the same way as $V$.

\textbf{Predictive Candidates.} Most of the existing ITM embedding models are only limited to considering using their own model to train candidates. However, when the model processes sample pairs that are more difficult to distinguish, i.e., intractable negative pairs, the effect will be greatly reduced. To enhance the cross-modal global and local correlation between image and sentence, inspired by active learning, we try to add predictive candidates \cite{eisenschtat2017linking} extracted from other semantic networks to train with the main candidates. Specifically, we extract the features of small batches of $V_p=\{v_{1},...,v_{a}\}$ and $S_p=\{s_{1},...,s_{b}\}$ ($V_p$ and $S_p$ are in the same form as $V$ and $S$) features predicted by the first round of other models. For I2T, we calculate the cosine similarity of each $v_a$ and $S_p$ pair and denote it as $c_{v_{a}}$. Similarly, for T2I, we calculate the cosine similarity of each $s_b$ and $V_p$ pair and denote it as $c_{s_{b}}$. Then we have:

\begin{equation}
M_{vs}=\{c_{v_{1}},...,c_{v_{a}}\}; M_{sv}=\{c_{s_{1}},...,c_{s_{b}}\}
\end{equation}

The sets $M_{vs} \in R^{b\times a}$ and $M_{sv} \in R^{a\times b}$ can be expressed as shown in Eq. (3). For each element $c_{v_{a}}$ and $c_{s_{b}}$, we sort the cosine similarity scores and obtain the positions of the intractable training sample pairs, and save them in $P_{vs}$ and $P_{sv}$. The formula is shown in (4):

\begin{equation}
P_{vs}=\{h^k_{v_{1}},...,h^k_{v_{a}}\}; P_{sv}=\{h^q_{s_{1}},...,h^q_{s_{b}}\}
\end{equation}
where $h^k_{v_{a}}$ represents the position of the top-$k$ I2T intractable sample pairs and $h^q_{s_{b}}$ represents the position of the top-$q$ T2I intractable sample pairs. Finally, we need to process the location information $P_{vs}$ and $P_{sv}$, and get the predictive candidate results. For $P_{vs}$ and $P_{sv}$, we choose the top candidates and get the auxiliary intractable negative text candidates $\overline{t}=\{\overline{t}_{1},...,\overline{t}_{a}\}$ and image candidates $\overline{v}=\{\overline{v}_{1},...,\overline{v}_{b}\}$ by index, respectively. We then obtain the positive text candidates $\overline{\overline{t}}=\{\overline{\overline{t}}_{1},...,\overline{\overline{t}}_{a}\}$ corresponding to $\overline{v}$, and the positive image candidates $\overline{\overline{v}}=\{\overline{\overline{v}}_{1},...,\overline{\overline{v}}_{b}\}$ corresponding to $\overline{t}$, respectively. $\overline{v}$, $\overline{t}$, $\overline{\overline{v}}$, $\overline{\overline{t}}$ are the final predictive candidates that we got and their forms are the same as $v$ and $t$.

\subsection{Corpus Learning Module}\label{BB}

\textbf{Corpus Learning Module (CLM).} Here, we will focus on the Corpus Visual Learning Module (CVLM) and Corpus Textual Learning Module (CTLM), as shown in Fig. 2 and Fig. 3. For local-level regions visual features $V=\{v_{1},...,v_{o}\}, v_o \in R^d$, the self-attention method \cite{vaswani2017attention} can help us average feature $v_s=\frac{1}{o}\sum\nolimits_{i=1}^o v_{i}$ as the query to obtain global image representation $v_s$ by integrating image regions. To bridge the visual image and the corpus, we apply the inter-modal attention method to learn the global image representation $v_s$ and the corpus representation $Q$, to obtain the corpus image representation $v_c$. The specific formula is shown in (6).

\begin{figure}[htb]
    \centering
    \includegraphics[width=8cm]{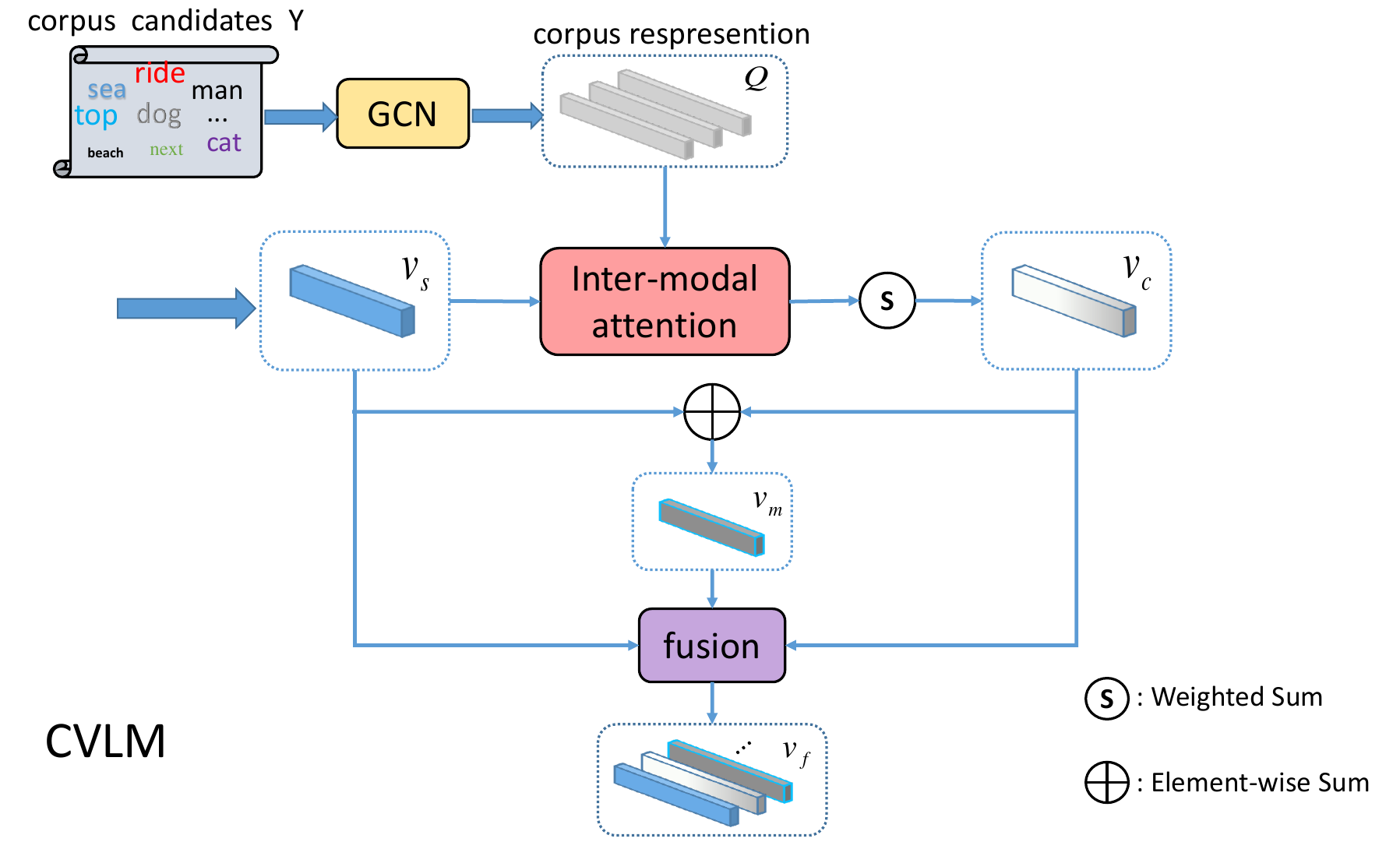}
    \caption{The architecture of CVLM.}
    \label{fig:CVLMframe}
\end{figure}

\begin{equation}
v_c=Q[f_a^{v}(v_s,Q)]=\sum\nolimits_{i=1}^z [f_a^{v}(v_s,Q)]q_i\atop f_a^{v}(v_s,Q)=\frac{exp(\lambda \hat{c}_i^v)}{\sum\nolimits_{i=1}^z exp(\lambda \hat{c}_i^v)}
\end{equation}
where $f_a^v(.)$ is inter modal attention function to calculate scores for $v_s$ in the corpus, $\lambda$ is the inverse temperature parameter of the softmax function, $\hat{c}_i^v=W^v_{\theta_1}v_s\cdot(W^v_{\theta_2}q_i)^T, W^v_{\theta_1}\in R^{d\times d}$ and $W^v_{\theta_2}\in R^{d\times d}$ are trainable projection parameters of $v_s$ and $q_i$, respectively. We take two vectors $v_s$ and $v_c$ as input and output a fused representation $v_m=u\cdot v_s+(1-u)\cdot v_c$. Finally, we stack the three features $v_s$, $v_c$, and $v_m$ together to obtain the final image feature $v_f$ by the fusion sub-module.

\begin{figure}[htb]
    \centering
    \includegraphics[width=8cm]{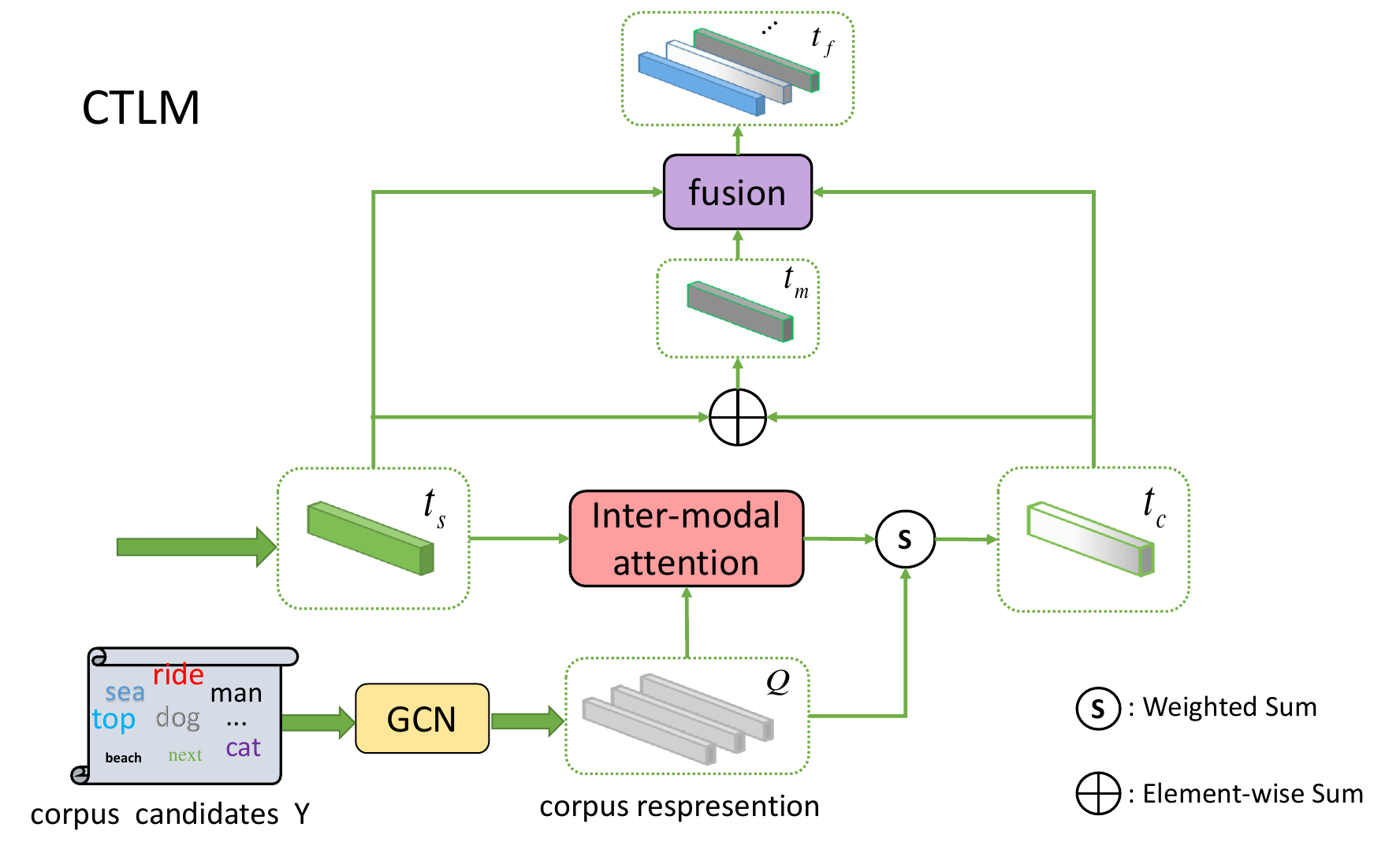}
    \caption{ The architecture of CTLM.}
    \label{fig:CTLMframe}
\end{figure}

Similarly, the global text representation $t_s$ is computed by the self-attention method from all the word features $T$. Its difference with CVLM is that CTLM adds the concept label $L \in R^{d\times 1}$ which is provided by Ref. \cite{wang2020consensus}. It can be used as the prior knowledge of the corpus text representation $t_c$. The specific formula is shown in Eq. (7):

\begin{equation}
t_c=Q[f_a^{t}(t_s,Q)]=\sum\nolimits_{j=1}^z [f_a^{t}(t_s,Q)]q_j\atop f_a^{t}(t_s,Q)=(1-\eta)\frac{exp(\lambda \hat{c}_j^t)}{\sum\nolimits_{j=1}^z exp(\lambda \hat{c}_j^t)}+\eta\frac{exp(\lambda \hat{c}_j^t)}{\sum\nolimits_{j=1}^z exp(\lambda \hat{c}_j^t)}
\end{equation}
where $f_a^t(,)$ and $\hat{c}_j^t$ correspond to $f_a^v(,)$ and $\hat{c}_i^v$ (have the same form), $\eta\in[0,1]$ is the ratio parameter of the prior label concept, $t_c$ is the corpus text representation. The practice and form of $t_m$ and $t_f$ are similar to $v_m$ and $v_f$.

\subsection{AHRL}\label{cc}
Our proposed loss measure AHRL ($L_{AHRL}$) can learn additional semantic discrimination information between other sample pairs in a diversified way. Our training is divided into two stages, and each stage has a different loss function. 

\textbf{Hierarchical Reinforcement Loss ($L_{HRL}$)}.
In the first phase, to improve the ability of the semantic model and to explore the fine-grained relationship in image-text pairs, the triplet loss function \cite{nam2017dual,wang2017deep} is the most direct and commonly used method. The formula is defined as:
\begin{equation}
{L_{trip}(v,t)=[\Delta_{1}-S_c(v,t)+S_c(v,t^{-})]_{+}\atop+[\Delta_{1}-S_c(v,t)+S_c(v^{-},t)]_{+}}
\end{equation}
here $(v,t)$ is the matched sample pair, $v^-$ and $t^-$ are negative sample pairs, $\Delta_{1}$ is the margin constraint, and $[\cdot] \equiv max(x,0)$, $S_c(\cdot,\cdot)$ represents the inner product, which is a method of calculating cosine similarity in our experiment.

Although the $L_{trip}$ loss can well motivate our model to narrow the distances between the anchors and the positives to predict a higher score. However, only using $L_{trip}$ loss in our model cannot achieve the best results in the experiment. The reason is that the datasets used for image and text matching is generally large-scale and the $L_{trip}$ loss is generally based on small batch anchor training, but it is not “harder” enough for intractable negative $v^-$ and $t^-$ training. Therefore, inspired by active learning, in the second phase of training, we add two unusual sample pairs of auxiliary training:

\begin{footnotesize}
\begin{equation}
L_{f_1}=[\Delta_{1}-S_c(v,t)+S_c(v,t^{-})]_{+}+[\Delta_{2}-S_c(v,t)+S_c(v,t^{-}_f)]_{+}+\atop[\Delta_{2}-S_c(v,t)+S_c(v^{-}_f,t^{-}_f)]_{+}
\end{equation}
\end{footnotesize}

\begin{footnotesize}
\begin{equation}
L_{f_2}=[\Delta_{1}-S_c(v,t)+S_c(v^{-},t)]_{+}+[\Delta_{2}-S_c(v,t)+S_c(v^{-}_f,t)]_{+}+\atop[\Delta_{2}-S_c(v,t)+S_c(v^{=}_f,t^{=}_f)]_{+}
\end{equation}
\end{footnotesize}here $\Delta_{1}$ is same as $\Delta_{2}$ is the margin constraint, $(v^{-}_f,t^{-}_f)$ are intractable negative pairs extracted from predictive candidates. $(v^{=}_f,t^{=}_f)$ are positive pairs corresponding to $(v^{-}_f,t^{-}_f)$, and they represent the sample pairs that are more difficult to distinguish.

In addition, we also follow Ref. \cite{chen2020adaptive} to add some penalty mechanisms so that the intractable negative pairs extracted by other high-quality models can be better combined with our model for training, the formula is defined as:
\begin{equation}
w_{1}=\tau-\frac{S_c(v,t^{-}_f)-S_c(v,t^{-})}{\mu}
;
w_{2}=\tau-\frac{S_c(v^{-}_f,t)-S_c(v^{-},t)}{\mu}
\end{equation}
Here $\tau$ and $\mu$ are penalty parameters. $w_1$ and $w_2$ are the penalty weights that can adaptively balance the training relationship of three negative pairs to improve the prediction score. So our final loss for the second round is expressed as $L_{HRL} = w_1L_{f_1}+w_2L_{f_2}$.

\textbf{Adaptive Reinforcement Margin.} 
Although Hierarchical Reinforcement Loss can find the ground-truth items corresponding to the provided image, some non-ground-truth items can also explain the provided image well, so the fixed margin scheme cannot greatly capture the semantic continuity.

Given an image $I$ as an anchor, due to Hierarchical Reinforcement Loss, some negative samples with correlation will be forced to move away from the anchor, although they can explain most of the semantics of the anchor. Inspired by \cite{biten2022image}, we adjust the margin of Hierarchical Reinforcement Loss with the following Eq. (11):
\begin{equation}
\Delta_{v}=\frac{\Phi(G_t,t)-\Phi(G_t,t^{-})}{\beta}
;
\Delta_{t}=\frac{\Phi(G_t,t)-\Phi(G_t,t^{=})}{\beta}
\end{equation}

here ${\Phi(.)}$ represents the captioning metric function CIDEr \cite{vedantam2015cider}, $G_t$ is the ground-truth caption set, $t^-$ and $t^=$ both are negative caption samples corresponding to $v$, $\beta$ is the temperature hyperparameter that controls margin scaling.

\textbf{AHRL.} We replace $\Delta_{1}$ of Eq. (8) and Eq. (9) with $\Delta_{v}$ and $\Delta_{t}$, then $L_{f_v}$ and $L_{f_t}$ are shown in Eq. (12) and Eq. (13):

\begin{footnotesize}
\begin{equation}
L_{f_v}=[\Delta_{v}-S_c(v,t)+S_c(v,t^{-})]_{+}+[\Delta_{2}-S_c(v,t)+S_c(v,t^{-}_f)]_{+}+\atop[\Delta_{2}-S_c(v,t)+S_c(v^{-}_f,t^{-}_f)]_{+}
\end{equation}
\end{footnotesize}

\begin{footnotesize}
\begin{equation}
L_{f_t}=[\Delta_{t}-S_c(v,t)+S_c(v^{-},t)]_{+}+[\Delta_{2}-S_c(v,t)+S_c(v^{-}_f,t)]_{+}+\atop[\Delta_{2}-S_c(v,t)+S_c(v^{=}_f,t^{=}_f)]_{+}
\end{equation}
\end{footnotesize}

Finally, our proposed AHRL is expressed as Eq. (14):

\begin{footnotesize}
\begin{equation}
L_{AHRL}=w_1L_{f_v}+w_2L_{f_t}.
\end{equation}
\end{footnotesize}

In general, our adaptive hierarchical reinforcement loss can not only better adaptively mine the intractable samples in the model but also add adaptive margins to each sample to better learn the semantics of query-related non-ground-truth items. Adaptive hierarchical reinforcement loss enables the model to have an excellent ability to distinguish positive and negative pairs and capture semantic continuity, no matter in the case of large-scale datasets or only small batches of training data.

\section{Experiments}
In this section, we not only verified the superiority of our model on two benchmark datasets but also elaborated on more training and experimental details.
\subsection{Datasets and Evaluation Protocols}
\textbf{Datasets.} Our experiments revolve around two large-scale ITM datasets: Flickr30k \cite{plummer2015flickr30k} and MSCOCO \cite{lin2014microsoft}. Flickr30k contains 31,783 images and each image comes with 5 unique annotation sentences. Following the strategy of \cite{faghri2017improving,karpathy2015deep}, we split the Flickr30k dataset into 1,000 test images, and 1,000 validation images, and the rest are training samples. Similarly, MSCOCO contains 123,287 pictures and each picture has 5 annotation sentences, 5,000 images are used as the validation set, 5,000 images are used as the test set, and the rest are used for training. 

\textbf{Evaluation Protocols.} Our model uses the most commonly used evaluation metric recall in image and text retrieval tasks to evaluate results, the recall at $K$ ($R@K$) metric represents the percentage of the correct results in the top-$k$ ranking results that are retrieved for a query source (an image or a text). Finally, for evaluation protocols, we not only adopted the $R@1$, $R@5$, and $R@10$ metrics but also adopted an additional metric $R@sum$ \cite{lee2018stacked}. $R@sum$ represents the sum of all metrics, and it can evaluate our model more comprehensively.
\subsection{Implementation Details}\label{SCM}
In this section, we will introduce the experimental details of our AMSPS model. For each image, the regions to be extracted are $o=36$ and get the 2048-dimensional feature for each region. For each sentence, the dimensions of the word vector and hidden layer are 300 and 1024, respectively. For the predictive candidates, we set the number of predictive image and text candidates as $a=400$ and $b=2000$, and the top candidate index list sizes to sample intractable negative image and text are set to $k=60$ and $q=300$ in Eq. (4). For the CLM, we set the dimension of the predictive score as 300 with smooth temperature $\lambda=10$ and $\eta=0.35$ in Eq. (6) and (7). For the loss function, we set $\Delta_{1}=0.2$ and $\Delta_{2}=0$ in Eq. (9) and Eq.(10) and $\tau=1.5$ and ${\mu}=0.3$ in Eq. (11). For the first phase training, the initial learning rate is 0.0002, and the second phase learning rate is reduced to 0.00002. 

\subsection{Performance Comparison}
In particular, We compare our AMSPS model with several well-known commonly used methods. Comparing these methods, we will prove the superiority and effectiveness of our model on the results of Flickr30k and MSCOCO.

\begin{table}[]
\small
\caption{Results by AMSPS and compared methods on MSCOCO.}
\begin{threeparttable}
\setlength{\tabcolsep}{1.2mm}{
\renewcommand\arraystretch{0.85}
\begin{tabular}{|clcccccccl|}
\hline
\multicolumn{10}{|c|}{MSCOCO 1$k$ test images}                                                                                                           \\ \hline
\multicolumn{2}{|c|}{}              & \multicolumn{3}{l|}{Sentence Retrieval} & \multicolumn{3}{l|}{Image Retrieval}    & \multicolumn{2}{l|}{}      \\
\multicolumn{2}{|c|}{Methods}        & $R@1$  & $R@5$  & \multicolumn{1}{c|}{$R@10$} & $R@1$  & $R@5$  & \multicolumn{1}{c|}{$R@10$} & \multicolumn{2}{c|}{$R@sum$}  \\ \hline
\multicolumn{2}{|c|}{VSE++ \cite{faghri2017improving}} & 64.6 & 90.0 & \multicolumn{1}{c|}{95.7} & 52.0 & 84.3 & \multicolumn{1}{c|}{92.0} & \multicolumn{2}{c|}{478.6} \\
\multicolumn{2}{|c|}{SCO \cite{huang2018learning}}  & 69.9 & 92.9 & \multicolumn{1}{c|}{97.5} & 56.7 & 87.5 & \multicolumn{1}{c|}{94.8} & \multicolumn{2}{c|}{499.3} \\
\multicolumn{2}{|c|}{CAMP \cite{wang2019camp}} & 72.3 & 94.8 & \multicolumn{1}{c|}{98.3} & 58.5 & 87.9 & \multicolumn{1}{c|}{95.0} & \multicolumn{2}{c|}{506.8} \\
\multicolumn{2}{|c|}{SCAN \cite{szegedy2016rethinking}}  & 72.7 & 94.8 & \multicolumn{1}{c|}{98.4} & 58.8 & 88.4 & \multicolumn{1}{c|}{94.8} & \multicolumn{2}{c|}{507.9} \\
 \multicolumn{2}{|c|}{VSRN* \cite{li2019visual}} & 74.0 & 94.3 & \multicolumn{1}{c|}{97.8} & 60.8 & 88.4 & \multicolumn{1}{c|}{94.1} & \multicolumn{2}{c|}{509.4} \\
\multicolumn{2}{|c|}{BFAN \cite{liu2019focus}} & 74.9 & 95.2 & \multicolumn{1}{c|}{98.3} & 59.4 & 88.4 & \multicolumn{1}{c|}{94.5} & \multicolumn{2}{c|}{510.7} \\
\multicolumn{2}{|c|}{MTFN \cite{karpathy2015deep}}  & 74.3 & 94.9 & \multicolumn{1}{c|}{97.9} & 60.1 & 89.1 & \multicolumn{1}{c|}{95.0} & \multicolumn{2}{c|}{511.3} \\
\multicolumn{2}{|c|}{CVSE \cite{wang2020consensus}} & 74.8 & 95.1 & \multicolumn{1}{c|}{98.3} & 59.9 & 89.4 & \multicolumn{1}{c|}{95.2} & \multicolumn{2}{c|}{512.7} \\
\multicolumn{2}{|c|}{GSMN \cite{liu2020graph}}  & 74.7 & 95.3 & \multicolumn{1}{c|}{98.2} & 60.3 & 88.5 & \multicolumn{1}{c|}{94.6} & \multicolumn{2}{c|}{511.6} \\ \hline
\multicolumn{2}{|c|}{ours} &
  \textbf{75.3} &
  \textbf{95.5} &
  \multicolumn{1}{c|}{\textbf{98.8}} &
  \textbf{61.4} &
  \textbf{90.3} &
  \multicolumn{1}{c|}{\textbf{96.1}} &
  \multicolumn{2}{c|}{\textbf{517.4}} \\ \hline
\end{tabular}}
\end{threeparttable}
\end{table}

\textbf{Comparisons on MSCOCO.} As shown in Table 1, on the MSCOCO dataset, our AMSPS method is better than these existing methods, with the best $R@1$=61.4\% for image retrieval and the best $R@sum$=517.4 when using the 1$k$ test images. It is worth noting that our model is improved based on CVSE, and the experimental results show that we have improved in all indicators, especially in image retrieval.

\begin{table}[]
\small
\caption{Results by AMSPS and compared methods on Flickr30k.}
\begin{threeparttable}
\setlength{\tabcolsep}{1.2mm}{
\renewcommand\arraystretch{0.85}
\begin{tabular}{|clcccccccl|}
\hline
\multicolumn{10}{|c|}{Flickr30k 5$k$ test images
}                                                                                                           \\ \hline
\multicolumn{2}{|c|}{}              & \multicolumn{3}{l|}{Sentence Retrieval} & \multicolumn{3}{l|}{Image Retrieval}    & \multicolumn{2}{l|}{}      \\
\multicolumn{2}{|c|}{Methods}        & $R@1$  & $R@5$  & \multicolumn{1}{c|}{$R@10$} & $R@1$  & $R@5$  & \multicolumn{1}{c|}{$R@10$} & \multicolumn{2}{c|}{$R@sum$}  \\ \hline
\multicolumn{2}{|c|}{VSE++ \cite{faghri2017improving}} & 52.9 & 80.5 & \multicolumn{1}{c|}{87.2} & 39.6 & 70.1 & \multicolumn{1}{c|}{79.5} & \multicolumn{2}{c|}{409.8
} \\
\multicolumn{2}{|c|}{SCO \cite{huang2018learning}}  & 55.5 & 82.0 & \multicolumn{1}{c|}{89.3} & 41.1 & 70.5 & \multicolumn{1}{c|}{80.1} & \multicolumn{2}{c|}{418.5
} \\
\multicolumn{2}{|c|}{SCAN \cite{szegedy2016rethinking}} & 67.4 & 90.3 & \multicolumn{1}{c|}{95.8} & 48.6 & 77.7 & \multicolumn{1}{c|}{85.2} & \multicolumn{2}{c|}{465
} \\
\multicolumn{2}{|c|}{MTFN \cite{karpathy2015deep}}  & 65.3 & 88.3 & \multicolumn{1}{c|}{93.3} & 52.0 & 80.1 & \multicolumn{1}{c|}{86.1} & \multicolumn{2}{c|}{465.1
} \\
\multicolumn{2}{|c|}{CAMP \cite{wang2019camp}} &  68.1 & 89.7 & \multicolumn{1}{c|}{95.2} & 51.5 & 77.1 & \multicolumn{1}{c|}{85.3} & \multicolumn{2}{c|}{466.9
} \\
\multicolumn{2}{|c|}{VSRN* \cite{li2019visual}} & 70.4 & 89.2 & \multicolumn{1}{c|}{93.7} & 53.0 & 77.9 & \multicolumn{1}{c|}{85.7} & \multicolumn{2}{c|}{469.9
} \\
\multicolumn{2}{|c|}{BFAN \cite{liu2019focus}}  & 68.1 & 91.4 & \multicolumn{1}{c|}{95.9} & 50.8 & 78.4 & \multicolumn{1}{c|}{85.8} & \multicolumn{2}{c|}{470.4
} \\
\multicolumn{2}{|c|}{CVSE \cite{wang2020consensus}} & 73.5 & 92.1 & \multicolumn{1}{c|}{95.8} & 52.9 & 80.4 & \multicolumn{1}{c|}{87.8} & \multicolumn{2}{c|}{482.5
} \\
\multicolumn{2}{|c|}{GSMN \cite{liu2020graph}}  & 72.6 & \textbf{93.5} & \multicolumn{1}{c|}{\textbf{96.8}} & 53.7 & 80.0 & \multicolumn{1}{c|}{87.0} & \multicolumn{2}{c|}{483.6} \\ \hline
\multicolumn{2}{|c|}{ours} &
  \textbf{73.9} &
  92.4 &
  \multicolumn{1}{c|}{96.3} &
  \textbf{54.1} &
  \textbf{82.0} &
  \multicolumn{1}{c|}{\textbf{88.9}} &
  \multicolumn{2}{c|}{\textbf{487.6}} \\ \hline
\end{tabular}}
\end{threeparttable}
\end{table}

\textbf{Comparisons on Flickr30k.} Table 2 shows the results on the Flickr30k dataset. Compared with the MSCOCO dataset, we find that our model has a greater improvement on Flickr30k with the best $R@5$=82.0\% for Image retrieval and the best $R@sum$=487.6. From this comparison and verification, we found that our model may have a more significant effect on a relatively small dataset.

\subsection{Ablation Study}
In this section, we will conduct several sets of ablation experiments to explore the impact of different components, and all these results will be based on the Flickr30k dataset. We will discuss the different results brought by different settings of the objective function and the improvement of hybrid retrieval in the testing.

\textbf{Different configurations of the loss function.} As shown in Table 3, We show different results from different configurations of the top list size of the intractable negative sample, $q$ and $k$ are the sizes of the intractable negative image list and text list, respectively (an image has 5 descriptions, so the size of $k$ is five times that of $q$). It can be seen from the table that the retrieval effect is better when $q$ is between 200 and 500 and $k$ is between 40 and 100. We finally chose $q$=300 and $k$=60 because it has the highest search score. When $q$=1000 and $k$=200, the intractable degree of the extracted samples decreases, but the effect of the model does not decrease much, indicating that the addition of the intractable samples greatly improves the generalization ability of the model.

\begin{table}[]
\footnotesize
\caption{The influence of the top list size of the intractable samples.}
\begin{threeparttable}
\setlength{\tabcolsep}{0.8mm}{
\renewcommand\arraystretch{0.85}
\begin{tabular}{|clcccccccl|}                               \hline
\multicolumn{2}{|c|}{}              & \multicolumn{3}{l|}{Sentence Retrieval} & \multicolumn{3}{l|}{Image Retrieval}    & \multicolumn{2}{l|}{}      \\
\multicolumn{2}{|c|}{Models}        & $R@1$  & $R@5$  & \multicolumn{1}{c|}{$R@10$} & $R@1$  & $R@5$  & \multicolumn{1}{c|}{$R@10$} & \multicolumn{2}{c|}{$R@sum$}  \\ \hline
\multicolumn{2}{|c|}{Base} & 72.5 & 91.3 & \multicolumn{1}{c|}{95.8} & 52.6 & 80.3 & \multicolumn{1}{c|}{87.2} & \multicolumn{2}{c|}{479.7

} \\
\multicolumn{2}{|c|}{Base+(k=40, q=200)
}  & 72.6  & 91.5 & \multicolumn{1}{c|}{\textbf{95.9}} & 53.8 & 81.8 & \multicolumn{1}{c|}{88.7} & \multicolumn{2}{c|}{484.3
} \\
\multicolumn{2}{|c|}{Base+(k=100, q=500)
} & \textbf{72.9} & \textbf{91.6} & \multicolumn{1}{c|}{\textbf{95.9}} & \textbf{54.0} & \textbf{81.9} & \multicolumn{1}{c|}{\textbf{88.8}} & \multicolumn{2}{c|}{\textbf{485.1}

} \\
\multicolumn{2}{|c|}{Base+(k=200, q=1000)}  & 72.8  & 91.5 & \multicolumn{1}{c|}{95.8} & 53.7 & 81.7 & \multicolumn{1}{c|}{88.6} & \multicolumn{2}{c|}{484.1

} \\
 \hline
\end{tabular}}
\end{threeparttable}
\end{table}

Table 4 shows the results obtained by configuring different penalty factors (from eq.(11)) of the loss function. From Table 4, we can see that it has a positive effect on the loss function.

\begin{table}[]
\footnotesize
\caption{Explore the influence of different penalty factors.}
\begin{threeparttable}
\setlength{\tabcolsep}{0.8mm}{
\renewcommand\arraystretch{0.85}
\begin{tabular}{|clcccccccl|}                               \hline
\multicolumn{2}{|c|}{}              & \multicolumn{3}{l|}{Sentence Retrieval} & \multicolumn{3}{l|}{Image Retrieval}    & \multicolumn{2}{l|}{}      \\
\multicolumn{2}{|c|}{Models}        & $R@1$  & $R@5$  & \multicolumn{1}{c|}{$R@10$} & $R@1$  & $R@5$  & \multicolumn{1}{c|}{$R@10$} & \multicolumn{2}{c|}{$R@sum$}  \\ \hline
\multicolumn{2}{|c|}{Base} & 72.5 & 91.3 & \multicolumn{1}{c|}{95.8} & 52.6 & 80.3 & \multicolumn{1}{c|}{87.2} & \multicolumn{2}{c|}{479.7

} \\
\multicolumn{2}{|c|}{Base+($\tau$ = 1.0, $\mu$ = 0.3)}
& 72.6 & 91.5 & \multicolumn{1}{c|}{95.9} & 53.8 & 81.8 & \multicolumn{1}{c|}{88.7} & \multicolumn{2}{c|}{484.3} \\
\multicolumn{2}{|c|}{Base+($\tau$  = 1.5, $\mu$ = 0.5)
} & 73.1 & 91.7 & \multicolumn{1}{c|}{95.8} & 53.9 & 81.8 & \multicolumn{1}{c|}{88.8} & \multicolumn{2}{c|}{485.1} \\
\multicolumn{2}{|c|}{Base+($\tau$  = 2.0, $\mu$ = 0.3)}  & 73.1  & \textbf{91.8} & \multicolumn{1}{c|}{\textbf{96.0}} & 53.9 & 81.9 & \multicolumn{1}{c|}{88.7} & \multicolumn{2}{c|}{485.4} \\
\multicolumn{2}{|c|}{Base+($\tau$  = 1.5, $\mu$ = 0.2)}  & 72.8  & 91.6 & \multicolumn{1}{c|}{95.9} & \textbf{54.3}  & \textbf{82.1} & \multicolumn{1}{c|}{\textbf{88.9}} & \multicolumn{2}{c|}{485.6} \\
\multicolumn{2}{|c|}{Base+($\tau$  = 1.5, $\mu$ = 0.3)}  & \textbf{73.2}  & \textbf{91.8} & \multicolumn{1}{c|}{95.9} & 54.1 &  82.0  & \multicolumn{1}{c|}{\textbf{88.9}} & \multicolumn{2}{c|}{\textbf{485.9}} \\
 \hline
\end{tabular}}
\end{threeparttable}
\end{table}

\textbf{The influence of local modules on the overall model.} Table 5 intuitively shows the impact of different local modules on the overall model. It is worth noting that although the Distributed Loss (DL) method can help the model improve the ability to identify intractable negative pairs and greatly increase the score, we found that the effect of sentence retrieval is not as good as image retrieval. Therefore, we chose to add the hybrid retrieval method in the test phase to re-ranking and only apply it to sentence retrieval to improve the I2T score of the model.

\begin{table}[]
\footnotesize
\caption{Explore the influence of local modules on the overall model.}
\begin{threeparttable}
\setlength{\tabcolsep}{1.43mm}{
\renewcommand\arraystretch{0.85}
\begin{tabular}{|clcccccccl|}                               \hline
\multicolumn{2}{|c|}{}              & \multicolumn{3}{l|}{Sentence Retrieval} & \multicolumn{3}{l|}{Image Retrieval}    & \multicolumn{2}{l|}{}      \\
\multicolumn{2}{|c|}{Models}        & $R@1$  & $R@5$  & \multicolumn{1}{c|}{$R@10$} & $R@1$  & $R@5$  & \multicolumn{1}{c|}{$R@10$} & \multicolumn{2}{c|}{$R@sum$}  \\ \hline
\multicolumn{2}{|c|}{Base} & 72.5 & 91.3 & \multicolumn{1}{c|}{95.8} & 52.6 & 80.3 & \multicolumn{1}{c|}{87.2} & \multicolumn{2}{c|}{479.7

} \\
\multicolumn{2}{|c|}{Base+DL
}  & 73.2  & 91.8 & \multicolumn{1}{c|}{95.9} & \textbf{54.1} & \textbf{82.0}  & \multicolumn{1}{c|}{\textbf{88.9}} & \multicolumn{2}{c|}{485.9

} \\
\multicolumn{2}{|c|}{Base+HR(I2T)

} & 73.0 & 91.7 & \multicolumn{1}{c|}{96.1} & 52.6 & 80.3 & \multicolumn{1}{c|}{87.2} & \multicolumn{2}{c|}{480.9

} \\
\multicolumn{2}{|c|} {ours
}  & \textbf{73.9}  & \textbf{92.4} & \multicolumn{1}{c|}{\textbf{96.3}} & \textbf{54.1} & \textbf{82.0} & \multicolumn{1}{c|}{\textbf{88.9}} & \multicolumn{2}{c|}{\textbf{487.6}

} \\
 \hline
\end{tabular}}
\end{threeparttable}
\end{table}

\subsection{Visualization and Analysis}
\begin{figure}[htb]
    \centering
    \includegraphics[width=7cm]{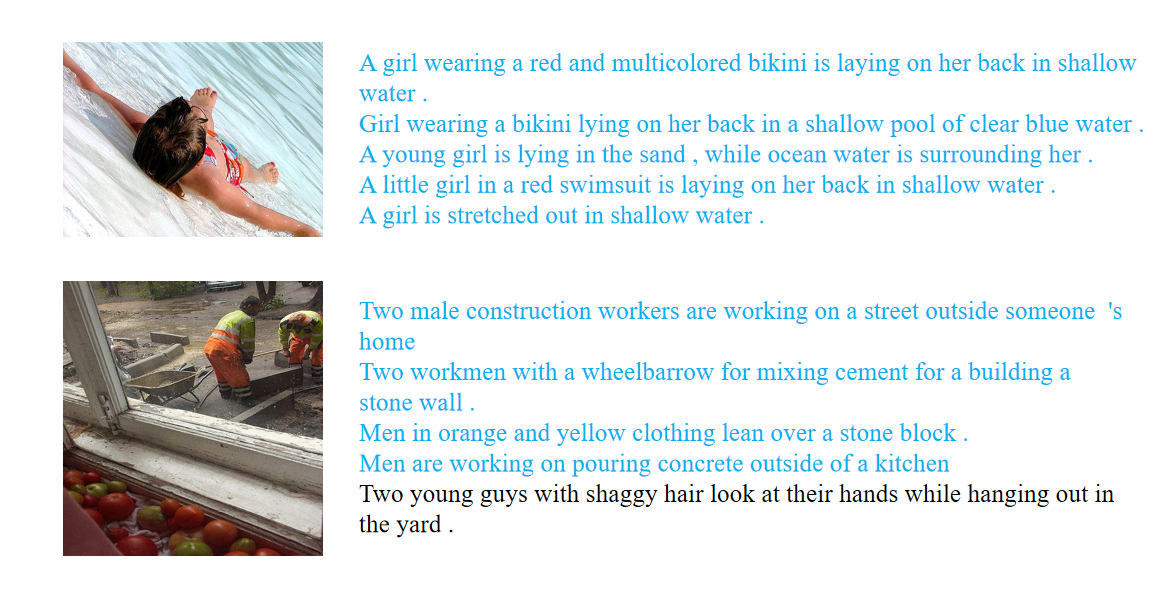}
    \caption{ An example represents the top five texts retrieved by our model through an image instance on Flickr30k. Blue font indicates matching sentences.}
    \label{fig:frame}
\end{figure}

\begin{figure}[htb]
    \centering
    \includegraphics[width=7cm]{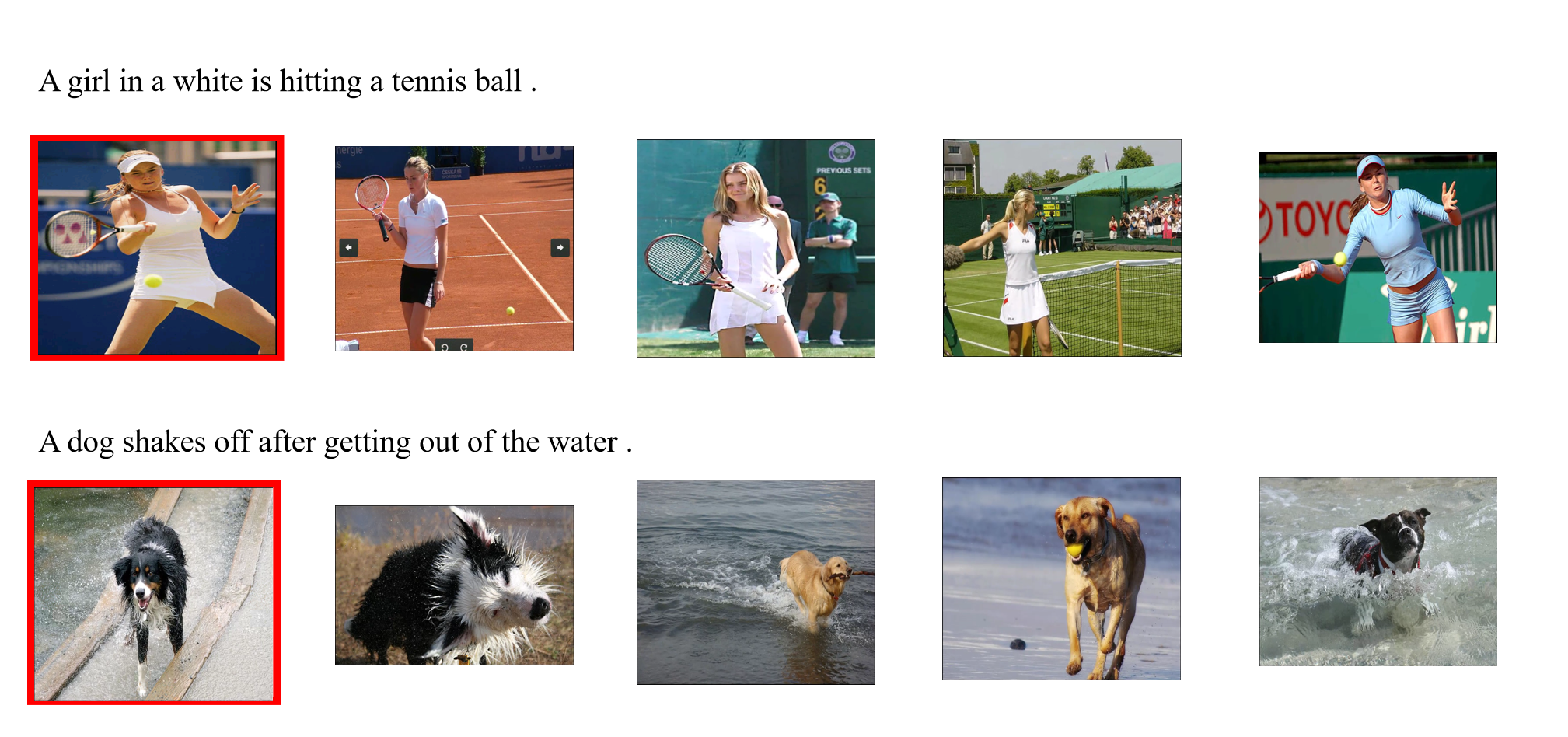}
    \caption{An example represents the top five images retrieved by our model through a text instance on Flickr30k. The image with the red border indicates the matching image.}
    \label{fig:frame}
\end{figure}

In order to show the ranking performance of our model more intuitively and clearly, we show examples of image and text retrieval in Fig.4 and Fig. 5, respectively. It is worth noting that an image has 5 corresponding descriptions, but a description has only one corresponding image. From Fig. 4, we know that our model has a very high accuracy rate in sentence retrieval. Almost all sentences matching the query can be retrieved. For example, the keywords ``girl", ``red" and ``water" in the first example have very similar semantics to the query image. The top$1$ image retrieved in Fig. 5 also corresponds to the sentence query. Intuitively, the rest of the image semantics is also very close. In the first four retrieved images, the model can match the image of a girl in white clothes holding a tennis racket, which is very similar in semantics. It can be seen that our model has been able to recognize the semantics of images or texts so as to ensure that the examples retrieved by the query are highly corresponding.

\section*{Conclusions}
Most of the existing commonsense learning models based on triplet loss are not strong in distinguishing between positive and negative pairs in the training phase. Moreover, in the test phase, the ranking score is not high and the generalization ability is poor. In this paper, we proposed an AMSPS model that divides the training into two stages and adds predicted candidates to enhance the ability to distinguish between positive and negative pairs. In addition, we apply the hybrid retrieval method to improve the poor reasoning ability of the semantic model in the test stage. Experiments under the baseline datasets show that compared with other models, our model can effectively improve the score ranking in a shorter training time, and the generalization ability of the model has been greatly improved.

\section*{Acknowledgment}

This work is supported in part by the Science and Technology Project of Guangdong Province (no.2021A1515011341), the Guangzhou Science and Technology Plan Project (no.202002030386), and the Guangdong Provincial Key Laboratory of Human Digital Twin (2022B1212010004).




\bibliographystyle{elsarticle-num} 
\bibliography{ITMbib}





\end{document}